\documentclass[letterpaper,10pt,conference]{ieeeconf}

\IEEEoverridecommandlockouts
\usepackage[T1]{fontenc}
\usepackage{times}
\ifdefined\XeTeXversion\else
  \usepackage{mathptmx}
\fi
\usepackage{cite}
\usepackage{amsmath,amssymb,amsfonts}
\usepackage{algorithmic}
\usepackage{graphicx}
\usepackage{textcomp}
\usepackage{xcolor}
\usepackage{etoolbox}

\makeatletter
\patchcmd{\@makecaption}
  {\footnotesize\scshape #2}
  {\footnotesize #2}
  {}{\PackageError{TRACE}{Could not set the table caption font}
    {Check the caption definition in ieeeconf.cls.}}
\newcommand{\tracefigurecaption}[2]{%
  \begingroup
  \def\@captype{figure}%
  \@makecaption{#1}{#2}%
  \endgroup}
\makeatother

\newcommand{\tracemethodmainimage}{%
  \begingroup
  \setlength{\unitlength}{\linewidth}%
  \begin{picture}(1,0.5159663866)
    \put(0,0){\includegraphics[width=\unitlength]{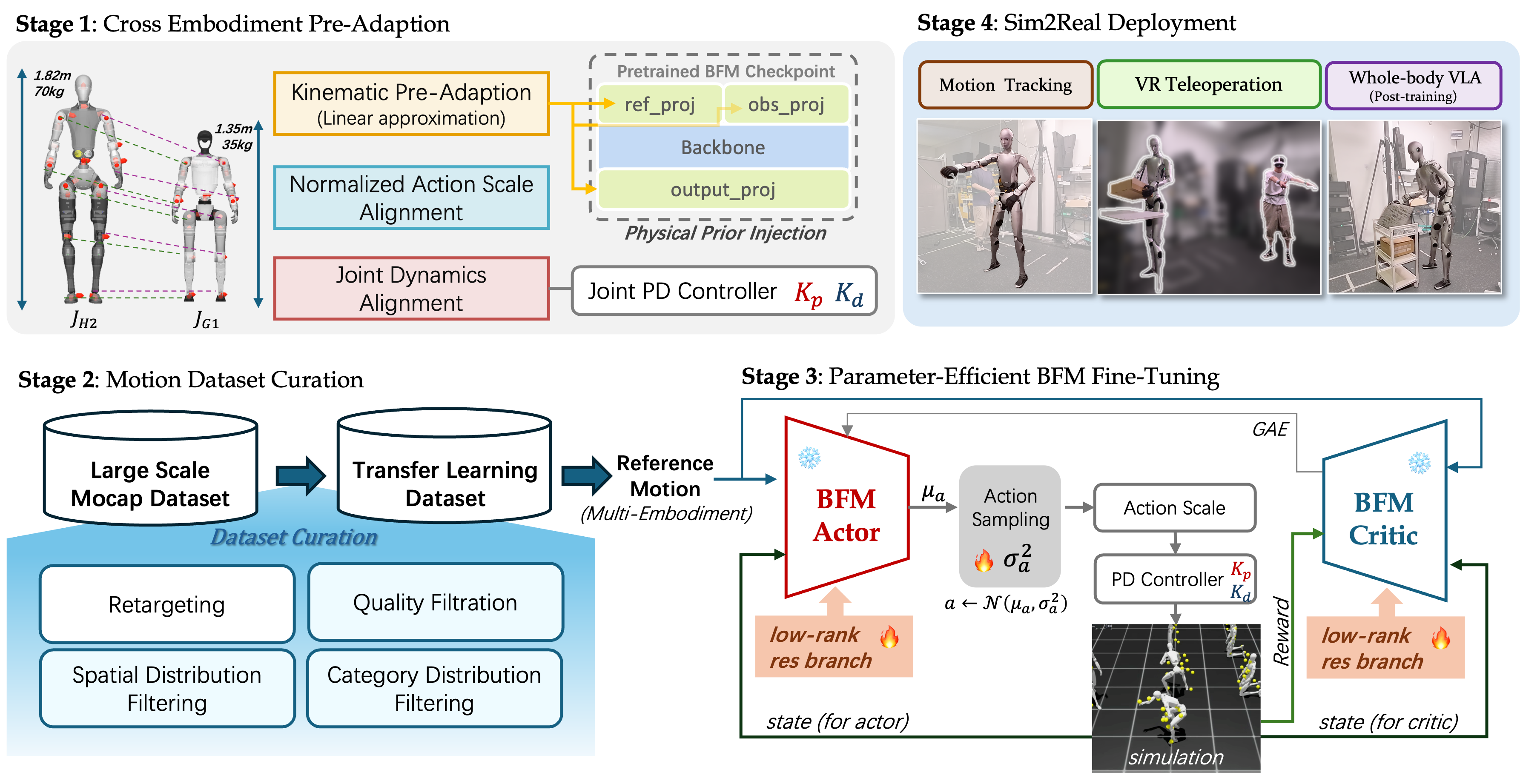}}%
    \put(0.1403361345,0.4266106443){%
      \color[RGB]{242,242,242}%
      \rule{0.0263305322\unitlength}{0.0084033613\unitlength}}%
    \put(0.1414565826,0.4280112045){%
      \resizebox{0.0246498599\unitlength}{0.0067226891\unitlength}{%
        \sffamily\bfseries\itshape 1.32m}}%
  \end{picture}%
  \endgroup}

\graphicspath{{imgs/}
  {00_Abstract/imgs/}
  {01_Introduction/imgs/}
  {02_RelatedWork/imgs/}
  {03_Method/imgs/}
  {04_Experiments/imgs/}
  {05_Conclusion/imgs/}
  {06_Discussion/imgs/}
  {99_Reference/imgs/}}

\def\BibTeX{{\rm B\kern-.05em{\sc i\kern-.025em b}\kern-.08em
  T\kern-.1667em\lower.7ex\hbox{E}\kern-.125emX}}

\usepackage[hidelinks]{hyperref}
\begin{document}

\title{\LARGE\bfseries
TRACE: Efficient Transfer Learning for Neural Whole-Body Control Across
Humanoid Embodiments}

\author{%
\authorblockN{Qingyu Lan$^{1}$, Zhihao Tian$^{1}$, Zifan Wang$^{1}$,
and Junwei Liang$^{1,2,\dagger}$}
\authorblockA{%
$^{1}$The Hong Kong University of Science and Technology (Guangzhou)\\
$^{2}$The Hong Kong University of Science and Technology\\
$^{\dagger}$Corresponding author\\[0.5ex]
Project page: \url{https://bfm-trace.github.io}}
}

\IEEEaftertitletext{%
  \vspace{-\baselineskip}
  \begin{center}
    \begin{minipage}{\textwidth}
      \centering
      \includegraphics[width=0.95\linewidth]{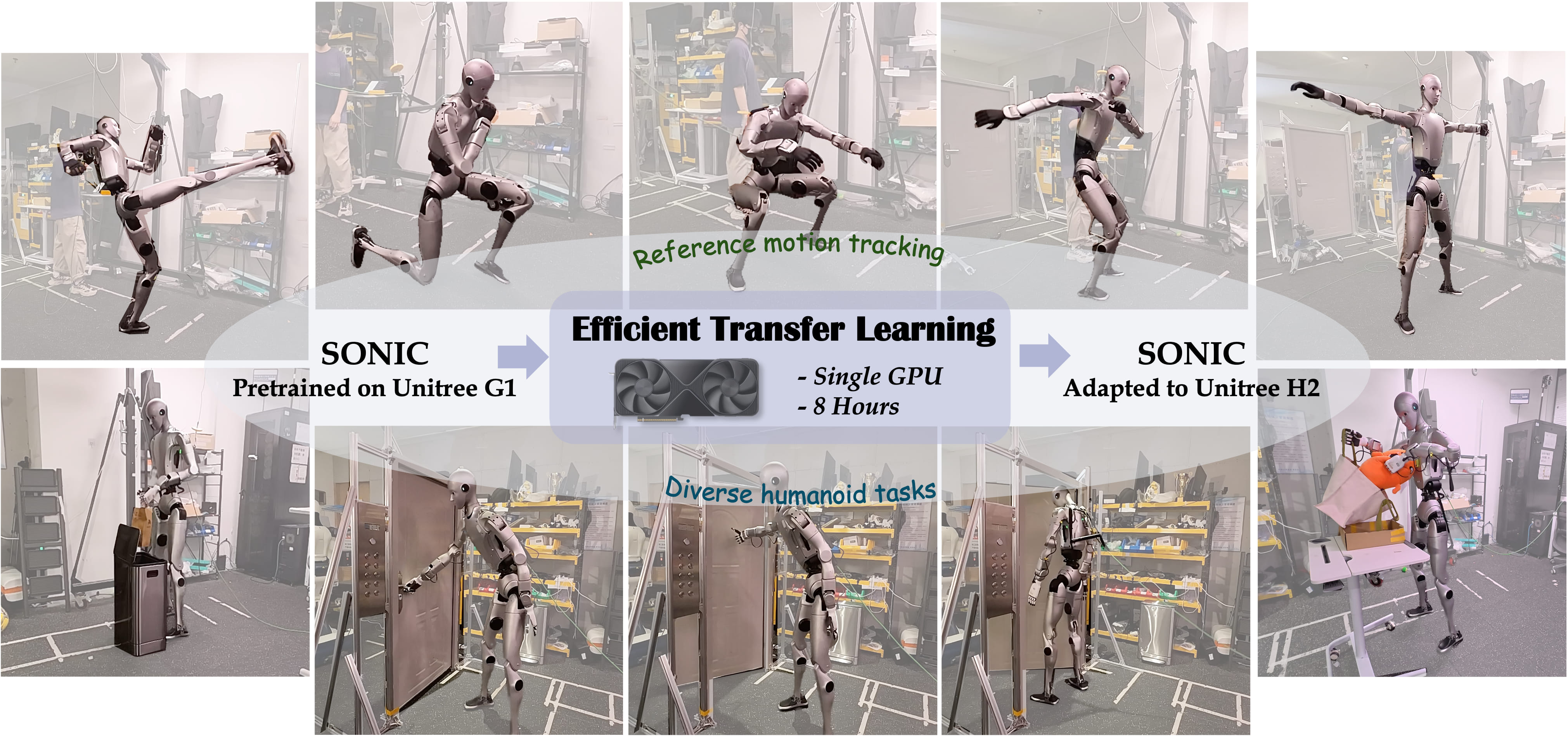}
      \setcounter{figure}{0}
      \refstepcounter{figure}\label{fig:trace-overview}
      \tracefigurecaption{Fig.~\thefigure}{TRACE enables efficient
      cross-embodiment transfer of humanoid behavior foundation models.
      We adapt the pretrained SONIC controller from the compact Unitree G1
      to the full-scale Unitree H2 in approximately eight hours of training on
      a single GPU. The resulting controller supports whole-body motion
      tracking and diverse downstream tasks.}
    \end{minipage}
  \end{center}
}

\maketitle
\thispagestyle{empty}
\pagestyle{empty}

\begin{abstract}
Humanoid behavior foundation models (BFMs) enable diverse whole-body control,
but adapting a pretrained controller to a new embodiment remains challenging
because of kinematic and dynamic mismatch.
We present TRACE, a framework for efficient cross-embodiment adaptation that
preserves pretrained motion representations while specializing feedback control
to the target robot.
TRACE first injects kinematic priors into input and output projection layers
and aligns nominal joint dynamics to reduce embodiment mismatch
and support stable fine-tuning.
It then combines encoder distillation with low-rank reinforcement learning on a
compact, quality-screened motion set, using a two-phase curriculum to establish
stable tracking before refining difficult motions.
We apply TRACE to transfer SONIC from Unitree G1 to Unitree H2 using 10 million
motion frames and approximately eight hours of training on a single RTX PRO 6000
GPU, at less than 0.1\% of the GPU-hour cost of training SONIC from
scratch.\begingroup
\renewcommand{\thefootnote}{\fnsymbol{footnote}}%
\footnote[1]{SONIC's reported training cost of 21,000 GPU-hours~\cite{luo2025sonic}
serves as the reference. TRACE's cost covers only adaptation from the released
SONIC checkpoint, excluding source pretraining. GPU-hours are not normalized
across hardware.}%
\endgroup{}
The adapted controller achieves 96.08\% success on 67,953 evaluation motions
under a height-scaled tracking criterion.
Real-robot demonstrations span dynamic motion tracking and whole-body
teleoperation, with collected demonstrations supporting downstream
vision\textendash{}language\textendash{}action policy post-training.
These results demonstrate practical reuse of a pretrained humanoid controller
across robots with different body scales and dynamics.
\end{abstract}

\section{Introduction}
\label{sec:introduction}

Humanoid behavior foundation models (BFMs) aim to learn a single whole-body
controller for diverse behaviors rather than behavior-specific
policies~\cite{yuan2025bfmsurvey}. Training on large motion datasets with a
shared tracking objective makes this approach scalable. Reference motions also
provide a compact representation and command interface for higher-level models
to specify whole-body behavior.
SONIC exemplifies this paradigm: trained on over 100 million motion frames,
a single Unitree G1 policy supports tracking unseen motions, VR teleoperation,
and vision\textendash{}language\textendash{}action (VLA) control~\cite{luo2025sonic}.

Although a BFM trained for a single embodiment can generalize across behaviors,
cross-embodiment transfer requires addressing two principal mismatches. First,
kinematic mismatch stems from differences in the number and arrangement of
actuators and sensors, as well as in link geometry, altering the dimensionality
and physical semantics of the BFM's observation and action spaces. Second,
dynamics mismatch may persist even after kinematic correspondence is established.
Differences in link masses and other inertial properties, actuator dynamics, and
contact dynamics may induce different state transitions under identical commands
and feedback, leading to different closed-loop responses that can affect
stability. Neither mismatch is readily resolved by modifying only the BFM's
inputs and outputs. Yet training a dedicated BFM for each embodiment is
prohibitively expensive, particularly as robotic platforms evolve rapidly.

Recent multi-embodiment approaches jointly train shared controllers across
multiple humanoid embodiments or randomized morphologies, often using unified
observation\textendash{}action interfaces and embodiment-aware
architectures~\cite{lin2025hzero,xue2026xhugwbc,peng2026eagle}. Their goal is
zero-shot generalization or few-shot adaptation to new
embodiments~\cite{lin2025hzero,xue2026xhugwbc}. However, broader embodiment
coverage increases training cost, while control performance on out-of-distribution
embodiments remains uncertain~\cite{lin2025hzero,xue2026xhugwbc}. Shared policies
may also fall short of embodiment-specific policies in control accuracy and
stability~\cite{xue2026xhugwbc,peng2026eagle}. Recent work also enables efficient
transfer of pretrained BFMs by combining joint-level kinematic alignment with low-rank
adaptation of dynamics-sensitive modules~\cite{yang2026any2any}. However, this
formulation treats the post-alignment gap as dynamical, without an explicit
objective for aligning motion latents across embodiments. It also retains the
source motion-sampling and optimization settings, leaving transfer-specific data
selection and staged control refinement outside its scope.

To this end, we introduce TRACE, an efficient
\textbf{TR}ansfer learning framework \textbf{AC}ross humanoid
\textbf{E}mbodiments for BFMs. TRACE reuses pretrained motion representations
while specializing feedback control for the target embodiment. We first inject
kinematic priors into the BFM's input and output projection layers for
kinematic pre-alignment, and further reduce the initial embodiment mismatch
through joint-command normalization and dynamic-response alignment. We also
introduce a targeted dataset construction
strategy that selects a compact transfer set from quality-screened motions,
balancing motion categories and promoting dispersion in the command feature
space. Using this set, we separate representation alignment from control
optimization during transfer training: encoder
distillation anchors target motion encodings to the pretrained latent space,
while reinforcement learning adapts the control decoder to target dynamics,
both through low-rank updates. To accelerate convergence and improve control
quality, we organize this training into two phases: first establishing stable
tracking across the selected motions, then emphasizing difficult motions to
improve tracking and contact accuracy.

We use TRACE to transfer the pretrained SONIC controller from Unitree G1 to
Unitree H2. The two embodiments differ in body scale, mass distribution, joint
geometry, and actuation. The transfer training completes
in approximately eight hours on a single RTX PRO 6000 GPU, at a fraction of
SONIC's 21k GPU-hour pretraining cost~\cite{luo2025sonic} and with improved
transfer efficiency over prior work~\cite{yang2026any2any}.
Quantitative evaluation
on a common motion-tracking benchmark validates the control quality achieved
after transfer, while ablation studies quantify the contributions of key
components. On the physical Unitree H2, we further demonstrate demanding motion
tracking and collect whole-body demonstrations through VR teleoperation. These
demonstrations are then used to post-train a VLA model for autonomous task
execution. These results show that TRACE enables efficient cross-embodiment BFM
transfer and provides a reusable controller for real-world tasks and downstream
policy learning.

The main contributions of this work are threefold:
\begin{itemize}
  \item We present TRACE, an efficient cross-embodiment transfer learning
  framework for BFMs that reuses pretrained motion representations while
  specializing feedback control for a new humanoid. Its core design combines
  kinematic and dynamic-response pre-alignment with two-phase low-rank training
  that decouples reference-encoder distillation from control optimization.

  \item We introduce a targeted motion-selection strategy that balances
  motion categories and promotes dispersion in the command feature space after
  quality screening. The resulting compact transfer set improves motion-tracking
  success under a fixed data budget.

  \item We achieve low-cost transfer of SONIC from Unitree G1 to the
  substantially different Unitree H2, with 96.08\% tracking success on the PHUMA
  benchmark. On the physical H2, the transferred controller supports demanding
  motion tracking and VR-based whole-body demonstration collection, with the
  collected data enabling VLA post-training for autonomous tasks.
\end{itemize}

\ifdefined\traceMethodMainFigureQueued\else
\global\let\traceMethodMainFigureQueued\relax
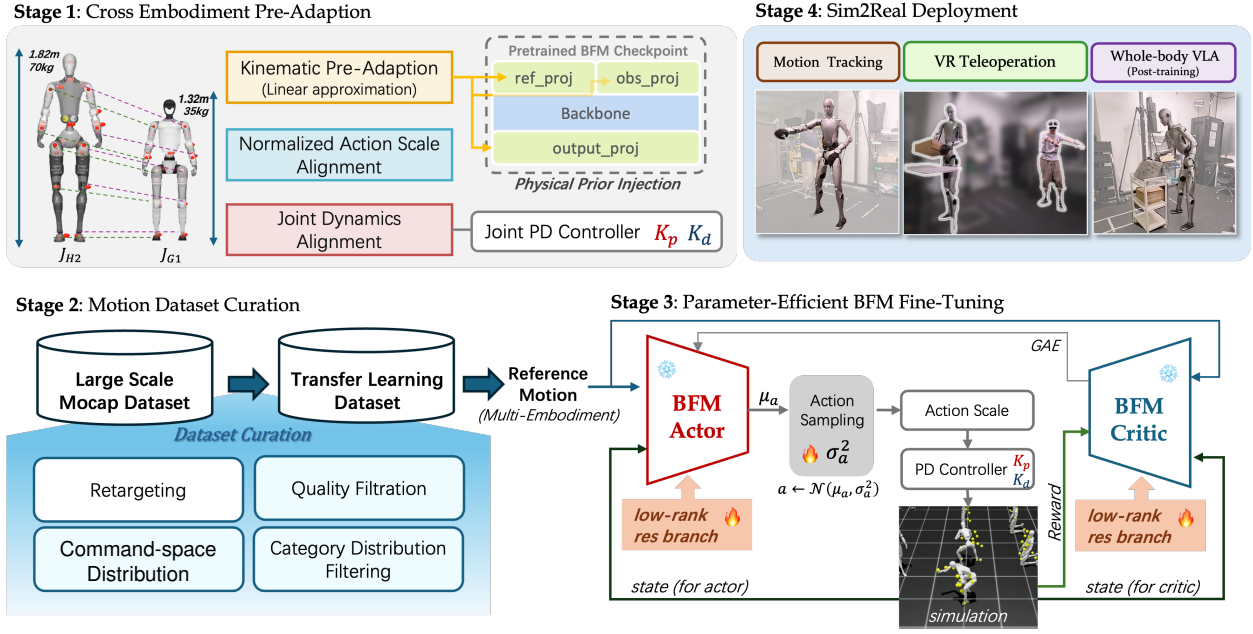
\begin{figure*}[!t]
  \centering
  \resizebox{0.93\textwidth}{!}{
\begingroup
\setlength{\unitlength}{\linewidth}%
\begin{picture}(1,0.5159663866)
  \put(0,0){\tracemethodmainimage}%
  \put(0.0420168067,0.0411764706){%
    \color[RGB]{238,251,255}%
    \rule{0.1372549020\unitlength}{0.0406162465\unitlength}}%
  \put(0.1098039216,0.0613445378){%
    \makebox(0,0){%
      \color{black}%
      \resizebox{0.1254901961\unitlength}{!}{%
        \sffamily\mdseries\shortstack{Command-space\\Distribution}}}}%
\end{picture}%
\endgroup}
  \caption{Overview of TRACE.
  \emph{Stage 1 (top left):} projection layers are inserted at the reference,
  observation, and action interfaces to inject kinematic priors; action scales
  are calibrated and nominal joint responses are aligned.
  \emph{Stage 2 (bottom left):} retargeting, quality screening, category balancing,
  and command-space selection form a compact motion dataset.
  \emph{Stage 3 (bottom right):} with base weights frozen within each training
  phase, low-rank residual branches adapt the target Robot reference encoder through latent and token
  distillation, and the Robot Control Decoder through PPO in target simulation
  (Fig.~\ref{fig:sonic-lora-details}).
  \emph{Stage 4 (top right):} the transferred controller supports diverse robot
  tasks, including reference-motion tracking, VR teleoperation, and downstream
  VLA-based loco-manipulation.}
  \label{fig:method-main}
\end{figure*}
\fi

\section{Related Work}
\label{sec:related-work}

\paragraph{Humanoid Behavior Foundation Model}
Humanoid BFMs build on reference-motion tracking in physics-based character
control. DeepMimic learned robust, physically plausible policies with dense
imitation rewards over prescribed clips; later work extended this to diverse
motions, dynamic and balance-intensive skills, and adaptation to real
humanoids~\cite{peng2018deepmimic,liao2026beyondmimic,he2025asap,zhang2025hub}.
Early generalists used broader motion libraries or unified command interfaces
within a single policy for universal tracking, teleoperation, and multimode
whole-body control~\cite{luo2023phc,he2025omnih2o,he2025hover}, extending beyond
individual references or small skill sets without jointly scaling motion data,
policy capacity, and distributed compute. Complementary unsupervised
reinforcement learning prioritizes stability and safety over fine-grained
tracking~\cite{li2026bfmzero}. SONIC established foundation-scale
reference-conditioned control by jointly scaling data, policy capacity, and
compute; trained on over 100 million motion frames, it generalizes to
unseen motions~\cite{luo2025sonic}.
Later work explores data and model scaling with billion-frame corpora and causal
Transformers~\cite{qi2026humanoidgpt}, and capacity scaling via conditional routing
in sparsely activated mixture-of-experts
Transformers~\cite{chen2026holomotion}. Existing BFMs remain coupled to their
training embodiment's state\textendash{}action interface and closed-loop dynamics.

\paragraph{Multi-Embodiment Generalist Control}
Multi-embodiment control assumes shared control-relevant physical regularities
across robots with different kinematics and dynamics. Early shared policies
encoded embodiment and pooled cross-design experience, enabling transfer to unseen
structures within a morphology space~\cite{gupta2022metamorph}.
Legged-robot methods combine morphology-agnostic diffusion with shared residual
adaptation~\cite{yang2025multiloco}, or large-scale training on procedurally
generated bodies with long-context test-time adaptation~\cite{liu2025locoformer}.
Recent humanoid systems combine unified state\textendash{}action semantics,
physically consistent morphology variation, embodiment- and topology-aware
policies, and generalist\textendash{}specialist refinement for zero- or few-shot
high-dimensional whole-body control and efficient target specialization
~\cite{lin2025hzero,xue2026xhugwbc,peng2026eagle}. These methods amortize learning
across embodiments, but training cost and coverage depend on the training
distribution's breadth and physical validity. Joint tokenization of embodiment,
state, and action also supports robot co-design and cross-embodiment
control~\cite{ha2026transformertransformer}, but scalability to humanoid BFM
pretraining remains unverified.

\paragraph{Transfer Learning for Humanoid BFMs}
BFM transfer reuses pretrained motion priors rather than relearning control.
Same-embodiment approaches typically add residual or delta policies to pretrained
trackers to adapt to new motion and task distributions or improve execution
accuracy~\cite{zeng2025bfm,wang2026fast,zhao2025resmimic,wang2026omnixtreme}.
With fixed kinematics and observation\textendash{}action interfaces, they do not
directly address embodiment transfer. Earlier cross-embodiment studies progressively optimized
policies along continuous morphology paths to narrow embodiment gaps, but required
repeated training and did not address parameter-efficient
transfer of large-scale humanoid BFMs~\cite{liu2022revolver,liu2024metaevolve}.
Recent work first validated parameter-efficient fine-tuning
(PEFT) for cross-embodiment humanoid BFM transfer~\cite{yang2026any2any}:
kinematic input/output alignment and LoRA~\cite{hu2022lora} fine-tuning of
dynamics-related modules enable low-cost transfer to multiple embodiments,
indicating that low-rank updates effectively capture this adaptation.
However, residual kinematic errors after joint-level alignment shift the command
distribution, while overemphasizing difficult motions early in fine-tuning erodes
pretrained priors; both slow convergence and reduce final tracking
accuracy.


\section{Methodology}
\label{sec:method}


Figure~\ref{fig:method-main} presents the four-stage TRACE framework.
Cross-embodiment pre-adaptation (Stage~1, Sec.~\ref{subsec:pre-adaptation})
injects kinematic priors into reference, observation, and action projections
and aligns action scales and nominal joint responses.
Motion dataset curation (Stage~2, Sec.~\ref{subsec:transfer-set}) combines
retargeting and quality screening with category balancing and command-space
selection. Parameter-efficient fine-tuning (Stage~3,
Sec.~\ref{subsec:policy-adaptation}) distills command latents and tokens from
the frozen source Robot reference encoder into its target counterpart, while
PPO adapts the control decoder to target dynamics.
The transferred controller is then deployed on the target robot (Stage~4),
supporting reference-motion tracking, VR teleoperation, and downstream
VLA-based loco-manipulation through a shared command interface.

\subsection{Problem Formulation}
\label{subsec:problem-formulation}

BFMs are pretrained by large-scale motion tracking, formulated as a discounted
MDP. For embodiment \(r\in\{s,t\}\), the policy
\(\pi_r(u_r\mid o_r,g_r)\) receives proprioception
\(o_r\) and a motion reference \(g_r\), and outputs normalized
joint commands \(u_r\). These define position targets for a PD controller;
rewards measure motion-tracking quality, and transitions follow the embodiment's
dynamics. We primarily focus on transferring SONIC-based BFMs from a half-scale
source humanoid \(s\) to a full-scale target humanoid \(t\). Our goal is to
accommodate differences in joint layouts and dynamics using limited target-side
training while retaining the pretrained motion representation.

\subsection{Cross-Embodiment Pre-Adaptation}
\label{subsec:pre-adaptation}

We align joint coordinates, action units, and nominal servo responses to reduce
initial tracking errors and stabilize early optimization while preserving the
source BFM's internal representation.

\paragraph{Kinematic prior injection}
To accommodate different joint layouts, we retain the pretrained model's source
joint-coordinate interface and inject kinematic priors into its reference,
proprioceptive, and action projection layers. We estimate the required linear maps
from paired retargeted poses, regularized by a Jacobian-based prior.

Let \(\Delta q_r=q_r-q_r^0\) denote joint offsets
from nominal pose \(q_r^0\), and let
\(Q_r\in\mathbb{R}^{n_r\times N}\) collect \(N\) paired joint-offset
samples for \(n_r\) joints. We construct the local prior
\(M_0=(J_t^0)^\dagger
D_{s\rightarrow t}J_s^0\) from limb-endpoint Jacobians.
Here \(J_r^0\) stacks the endpoint translational
Jacobians at the nominal pose in aligned limb-root frames,
\(D_{s\rightarrow t}\) scales each endpoint's three coordinates by
the corresponding target-to-source nominal root-to-endpoint length ratio, and
\(\dagger\) denotes the Moore--Penrose pseudoinverse. We fit the forward map by
\begin{equation}
  M_{s\rightarrow t}^{\star}
    =\arg\min_M \frac{1}{N}\|M Q_s-Q_t\|_F^2
      +\lambda_J\|M-M_0\|_F^2,
  \label{eq:jacobian-data-alignment}
\end{equation}
where \(\lambda_J>0\) controls the prior strength. The data term fits joint
coordination across poses, while the prior anchors directions poorly covered
by the data.

The same constant map is used for joint offsets and velocities:
\(\Delta q_t\approx M_{s\rightarrow t}^{\star}
\Delta q_s\) and
\(\dot{q}_t\approx M_{s\rightarrow t}^{\star}
\dot{q}_s\).
Exchanging \(s,t\) in the fit and prior gives a separately fitted reverse map.
We embed it in the full reference and proprioceptive input maps
\(x_s=P_i x_t+c_i\), \(i\in\{\mathrm{ref},\mathrm{obs}\}\), leaving
shared-coordinate channels unchanged. Here \(c_i\) contains nominal-pose
offsets, and \(P_{\mathrm{act}}\) denotes the normalized source-to-target action
map in \eqref{eq:target-action-interface}. Before fine-tuning, source projection
weights \(W^s\) and biases \(b^s\) become target parameters \(W^t,b^t\):
\begin{equation}
  \begin{alignedat}{2}
    & W_i^t = W_i^s P_i,
      && \qquad b_i^t = b_i^s+W_i^s c_i,\\[2pt]
    & W_{\mathrm{act}}^t = P_{\mathrm{act}}W_{\mathrm{act}}^s,
      && \qquad b_{\mathrm{act}}^t = P_{\mathrm{act}}b_{\mathrm{act}}^s.
  \end{alignedat}
  \label{eq:kinematic-prior-injection}
\end{equation}
Input projections therefore use right multiplication, whereas the action
projection uses left multiplication.

\paragraph{Action-scale alignment}
The kinematic map relates physical joint offsets, whereas the decoder outputs
normalized actions. Let \(S_r=\operatorname{diag}(s_r)\), where \(s_{r,j}>0\)
is the joint-angle offset per unit action. The target action
\(u_t=P_{\mathrm{act}}u_s\) and its PD position target are defined by
\begin{equation}
  \begin{aligned}
    P_{\mathrm{act}}&=S_t^{-1}M_{s\rightarrow t}^{\star}S_s,\\
    q_t^{\mathrm{cmd}}&=q_t^0+S_t\operatorname{clip}_{\mathcal{A}_t}(u_t),
  \end{aligned}
  \label{eq:target-action-interface}
\end{equation}
where clipping enforces the target normalized-action range \(\mathcal{A}_t\).

\paragraph{Joint dynamics alignment}
To reduce actuator-response mismatch, we initialize target PD gains using the
source controller's common nominal natural frequency \(\omega_n\) and damping
ratio \(\zeta\). Under a decoupled actuator model,
\begin{equation}
  K_{p,t,j}^{(0)}=I_{t,j}\omega_n^2,
  \qquad
  K_{d,t,j}^{(0)}
    =2\zeta\omega_n I_{t,j}-d_{t,j},
  \label{eq:pd-response-initialization}
\end{equation}
where \(I_{t,j}>0\) and \(d_{t,j}\ge0\) denote joint-side reflected actuator
inertia and passive damping at the nominal pose, with \(K_{d,t,j}^{(0)}>0\).
This actuator-level initialization excludes link and contact dynamics, which
are addressed by policy fine-tuning.

\subsection{Motion Dataset Curation}
\label{subsec:transfer-set}

Efficient transfer requires reliable motion references within a limited data
budget. We combine quality screening with category balancing and command-space
distribution shaping to construct the transfer set.

\paragraph{Retargeting and quality screening}
We retarget the motion corpus to both embodiments, obtaining paired references
for kinematic calibration and policy transfer. Quality screening rejects
discontinuities, sustained joint-position or
velocity-limit violations, unintended collisions, and implausible clip-level
acceleration or jerk. Contact checks detect excessive penetration, support-point
sliding, and missing intended support while preserving annotated object contacts.

\paragraph{Category balancing}
To avoid overrepresenting common motions, we divide a fixed frame budget equally
across ten metadata-derived categories (Fig.~\ref{fig:exp-data-strategy}(b))
in a fixed, quality-screened candidate pool.
If a category has fewer frames than its quota, we retain all its clips.
We redistribute the unused budget equally among categories with more frames
than their quotas.

\paragraph{Command-space distribution}
To promote command-space dispersion within each category, we summarize frozen
SONIC reference-encoder window features by their temporal means and standard
deviations, forming a 128-dimensional clip descriptor.
Let \(E_i\) be the mean squared component value of its standardized descriptor and
\(\widetilde E_i\) its counterpart after regularized covariance whitening.
Standardization parameters and covariance are estimated from the full valid
source pool. The clip score is
\begin{equation}
  s_i=\tfrac12\log\max(E_i,\epsilon)
      +\tfrac12\log\max(\widetilde E_i,\epsilon),
  \label{eq:motion-feature-score}
\end{equation}
where \(\epsilon=10^{-12}\) avoids taking the logarithm of zero.
\(E_i\) measures departure from the pool center in standardized coordinates,
while \(\widetilde E_i\) additionally accounts for feature correlations.
Their logarithms are equally weighted, and the resulting score guides selection
toward larger combined deviations within each category's frame budget.
Paired source--target references remain synchronized for distillation and tracking.

\subsection{Parameter-Efficient Fine-Tuning}
\label{subsec:policy-adaptation}

SONIC encodes references as Universal Tokens and combines them with
proprioception in a Robot Control Decoder. We adapt reference encoding and
feedback control separately to address residual representation and dynamics
mismatch.

\begin{figure}[t]
  \centering
  \includegraphics[width=\linewidth]{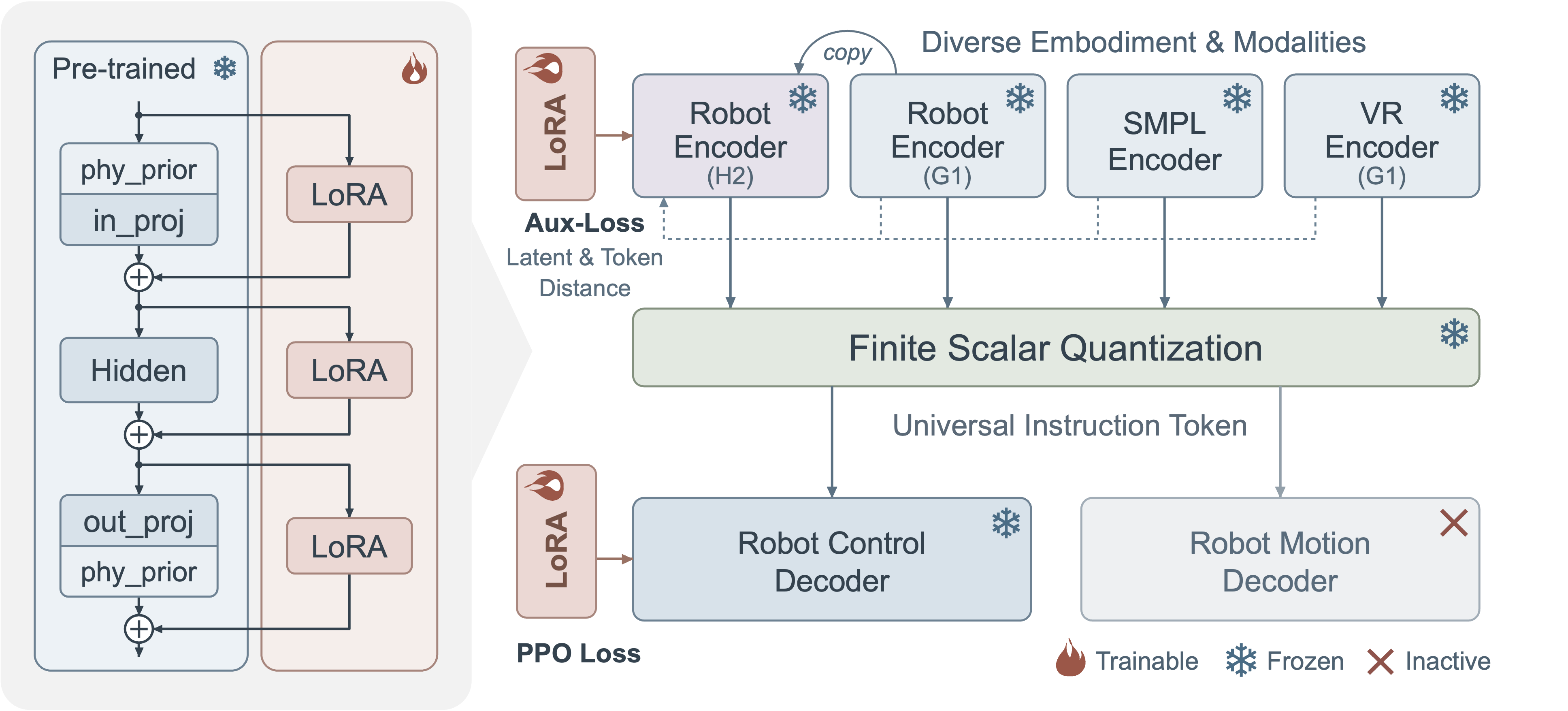}
  \caption{SONIC adaptation with kinematic priors and LoRA. Left: priors are
  folded into projection weights, and LoRA augments frozen layers. Right:
  the target Robot encoder uses latent and token distillation; the control
  decoder uses PPO. Source encoders and FSQ remain frozen.}
  \label{fig:sonic-lora-details}
\end{figure}

\paragraph{Model architecture and LoRA insertion}
As shown in Fig.~\ref{fig:sonic-lora-details}, we retain the source reference
motion encoders (Robot, Human, and Hybrid) as frozen branches. We initialize a
single target Robot reference encoder from its source counterpart with the
pre-adapted interface. All branches share the fixed finite scalar
quantization (FSQ) operator.

LoRA~\cite{hu2022lora} augments linear layers in the target Robot reference encoder
and Robot Control Decoder:
\begin{equation}
  W_{\mathrm{eff}}
    =W_b+\frac{\alpha}{r_{\mathrm{L}}}BA,
  \label{eq:lora-adaptation}
\end{equation}
where \(W_b\) is frozen within each training phase, \(\alpha\) is the scale,
and trainable factors \(A,B\) have inner dimension
\(r_{\mathrm{L}}\). LoRA branches in both the target Robot reference encoder and
control decoder use Kaiming-uniform initialization for \(A\) and
\(B=0\), preserving their pre-adapted mappings at insertion.
The critic uses LoRA in every linear layer except its trainable final
scalar projection. The Robot Motion Decoder remains frozen and excluded from
training because target-motion reconstruction is not required for feedback control.

\paragraph{Command latent alignment}
Preserving command-token semantics is a transfer requirement for reusing
multitask interfaces, including VR teleoperation, motion generators, and planners. Unconstrained
end-to-end RL risks semantic drift, whereas freezing the reference encoder
leaves residual kinematic-mapping errors that can shift token distributions
after nonlinear encoding and FSQ. We therefore fine-tune only the target Robot
reference encoder against its frozen source counterpart to maintain command
compatibility and support stable, data-efficient controller fine-tuning.
Let \(\mathbf{z}_s\) and \(\mathbf{z}_t\) denote their pre-FSQ command latents,
and \(\mathbf{c}_r=\mathcal{Q}(\mathbf{z}_r)\), \(r\in\{s,t\}\), their
quantized token vectors under the fixed FSQ operator \(\mathcal{Q}\).
We jointly align both representations by minimizing
\begin{equation}
  \mathcal{L}_{\mathrm{align}}
    =\ell_{\mathrm{SL1}}\!\left(\mathbf{z}_t,\mathbf{z}_s\right)
     +\ell_{\mathrm{SL1}}\!\left(\mathbf{c}_t,\mathbf{c}_s\right),
  \label{eq:decoupled-adaptation-objectives}
\end{equation}
where the same smooth-\(L_1\) loss \(\ell_{\mathrm{SL1}}\) measures both latent
and token distances.

\paragraph{Controller fine-tuning with PPO}
During rollouts, we sample the retained source reference branches and the
adapted target Robot branch
according to prescribed mixture proportions and use their quantized tokens as
commands. Each token remains paired with the corresponding target trajectory
for tracking-reward evaluation. PPO updates the Robot Control Decoder's LoRA
parameters and the action-sampling variance, with gradients stopped at the token
input; the critic receives the value loss.

\paragraph{Two-phase transfer learning curriculum}
Premature concentration of the training distribution on difficult motions can
disrupt the BFM's pretrained motion knowledge. We therefore adopt a two-phase
training curriculum within Stage~3. First, we sample from the balanced
transfer-set distribution and strengthen
existing global-position and end-effector tracking rewards to establish stable
target tracking. At the phase transition, we merge LoRA updates into the base
weights and reinitialize zero-output branches. Second, we adaptively
prioritize motion segments with frequent failures or large tracking errors
while retaining baseline sampling for coverage. We further strengthen
end-effector tracking rewards and introduce terrain-interaction and robustness
regularization to improve precision and disturbance rejection. Final LoRA
updates are merged for deployment.

\section{Experiments}
\label{sec:experiments}

\newcommand{\traceexptbd}{\textcolor{black!60}{\textsf{TBD}}}
We evaluate whether TRACE transfers broad tracking capability under a
limited adaptation budget and supports downstream use. Embodiment
differences and the availability of pretrained target policies and
compatible implementations limit direct comparisons with prior transfer methods.
We therefore emphasize training baselines and component ablations on the
target robot, using the source controller as a reference for tracking
quality in simulation. On hardware, where global tracking errors are
difficult to measure reliably, we assess challenging whole-body motions and
downstream task performance.

\edef\traceexpsavedtopnumber{\number\value{topnumber}}

\suppressfloats[t]
\begin{figure}[t]
  \centering
  \includegraphics[width=\linewidth]{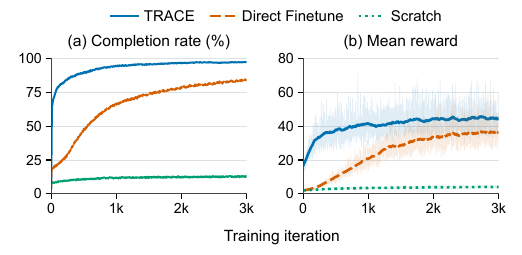}
  \caption{Convergence under a common motion-sampling distribution
  (difficulty-based oversampling disabled), with one run per method:
  (a)~completion rate; (b)~mean reward. TRACE converges within a small iteration budget;
  Direct Finetune requires substantially more training. Scratch fails to
  obtain a usable tracking policy within the 3k-iteration budget.}
  \label{fig:exp-training}
\end{figure}

\begin{figure}[t]
  \centering
  \includegraphics[width=\linewidth]{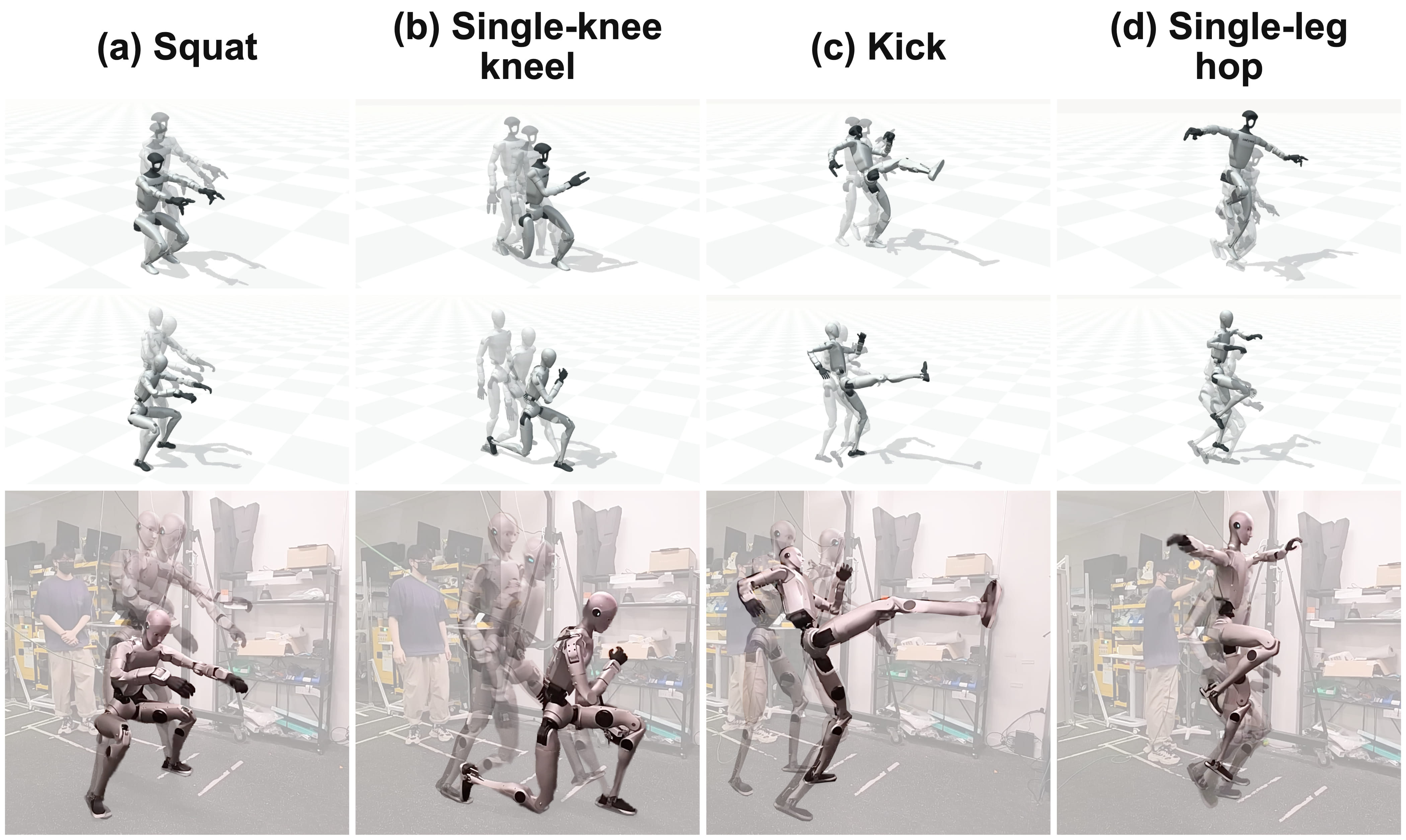}
  \caption{Motion tracking across embodiments and deployment settings.
  From top to bottom, rows show Unitree G1 simulation, Unitree H2 simulation,
  and H2 hardware execution. Columns show (a)~squatting, (b)~single-knee kneeling,
  (c)~kicking, and (d)~single-leg hopping.}
  \label{fig:exp-tracking}
\end{figure}

\subsection{Experimental Setup}
\label{subsec:exp-setup}

\paragraph{Training configuration}
We adapt the pretrained SONIC controller~\cite{luo2025sonic} from Unitree G1
to Unitree H2 in simulation, following Sec.~\ref{sec:method}. The adaptation
set contains 10M motion frames selected from BONES-SEED~\cite{bones_seed}.
Unless otherwise specified, LoRA modules use rank \(r_{\mathrm{L}}=32\).
Training uses 12,288 parallel
environments, a rollout horizon of 24 steps, and a batch size of 1,536.

\paragraph{Baselines}
We compare TRACE with \emph{Direct Finetune} (full-parameter fine-tuning) and
\emph{Scratch} (training without pretraining). Convergence is evaluated over
the first 3k iterations under a common motion-sampling distribution, with
difficulty-based oversampling disabled. We report completion rate, measured
as the episode-timeout fraction, and mean training reward.

\paragraph{Evaluation}
We evaluate motion tracking in simulation on the PHUMA benchmark~\cite{lee2025phuma},
using 67,953 motions excluded from H2 adaptation training. We evaluate the
late-training TRACE/H2 policy and re-evaluate the released SONIC/G1 policy
on the same motion set, with references retargeted to each robot. We report clip success
rate (SR), root-relative mean
body-position error \(E_{\mathrm{loc}}\) (MPJPE-L), velocity error
\(E_{\mathrm{vel}}\), and acceleration error \(E_{\mathrm{acc}}\).

\begin{table}[t]
  \centering
  \caption{Results on the PHUMA benchmark. SR uses
  body-size-adjusted thresholds: 0.250\,m (G1) and 0.3447\,m (H2).
  Errors are evaluated only on each policy's
  successful motions: $E_{\mathrm{loc}}$ in mm, $E_{\mathrm{vel}}$ in mm/frame,
  and $E_{\mathrm{acc}}$ in mm/frame$^2$, at 50\,Hz.}
  \label{tab:exp-transfer}
  \small\setlength{\tabcolsep}{1.5pt}
  \renewcommand{\arraystretch}{1.2}
  \begin{tabular*}{\linewidth}{@{\extracolsep{\fill}}cccccc@{}}
    \hline
    Method & Robot & SR (\%)$\uparrow$ & $E_{\mathrm{loc}}$ (mm)$\downarrow$ & $E_{\mathrm{vel}}\downarrow$ & $E_{\mathrm{acc}}\downarrow$ \\[1pt]
    \hline
    \noalign{\vskip 3pt}
    {\renewcommand{\arraystretch}{0.85}%
      \begin{tabular}[c]{@{}c@{}}SONIC\\{\footnotesize (Pre-trained)}\end{tabular}}
      & Unitree G1 & 97.59 & 27.81 & 5.157 & 1.390 \\[3pt]
    \hline
    \noalign{\vskip 3pt}
    TRACE (full)
      & Unitree H2 & 96.08 & 55.81 & 7.536 & 2.394 \\[3pt]
    {\footnotesize w/o-Pre-Adaption}
      & Unitree H2 & 87.29 & 72.88 & 8.033 & 2.474 \\[3pt]
    {\footnotesize w/o-Data-Strategy}
      & Unitree H2 & 93.80 & 54.01 & 7.853 & 2.558 \\[3pt]
    {\footnotesize w/o-Two-Phase-FT}
      & Unitree H2 & 95.64 & 62.47 & 8.115 & 2.553 \\[3pt]
    \hline
  \end{tabular*}
\end{table}

\begin{figure}[t]
  \centering
  \includegraphics[width=\linewidth]{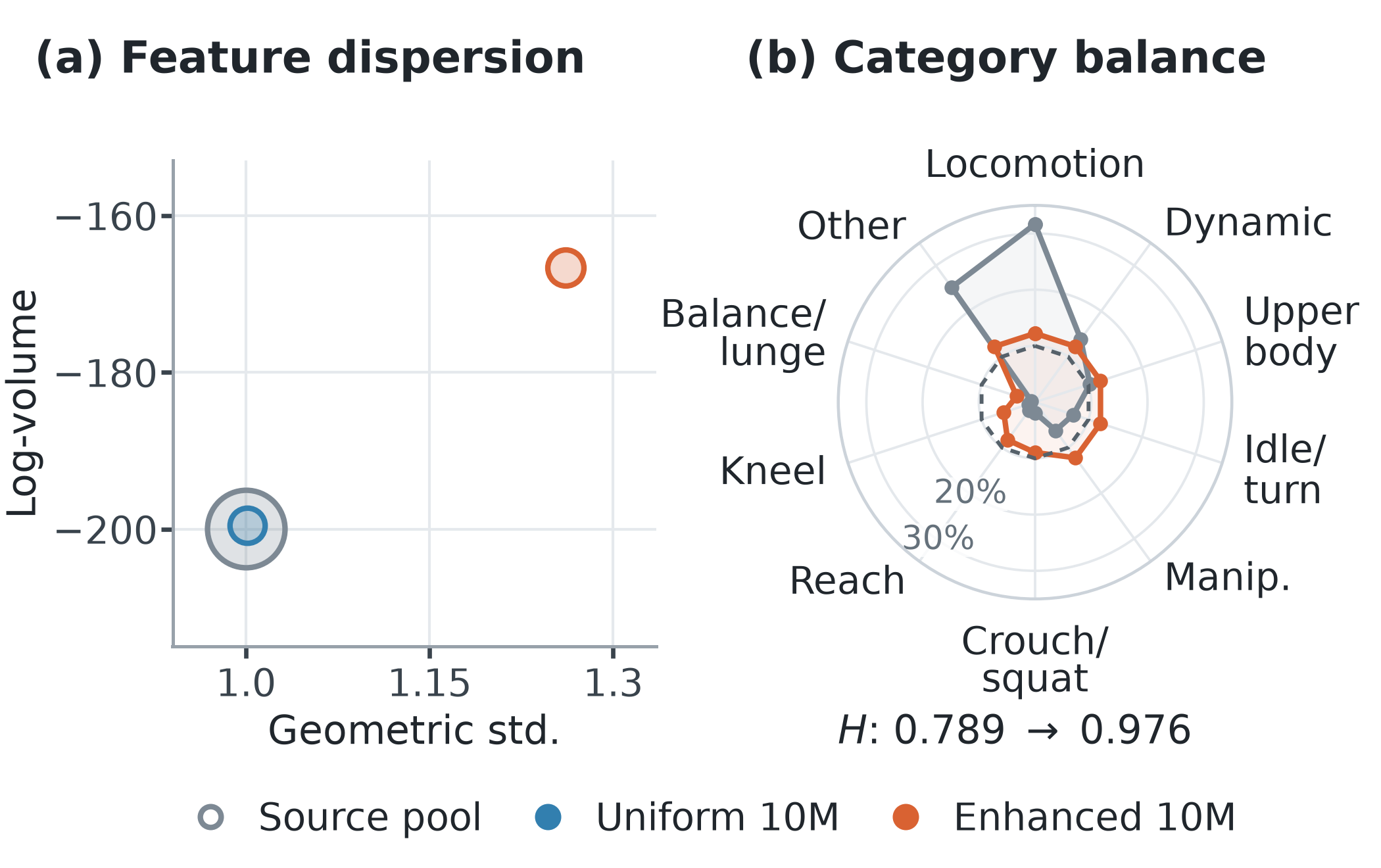}
  \caption{Data selection statistics. (a)~Geometric mean of feature standard
  deviations and regularized covariance log-volume of frozen SONIC descriptors
  under shared standardization.
  Coordinates average 20 draws of 5,000 clips; bubble areas indicate clip counts.
  (b)~Category frame shares; the dashed polygon marks equal shares (10\%).
  $H$ denotes normalized category entropy.}
  \label{fig:exp-data-strategy}
\end{figure}

\subsection{Transfer Efficiency and Control Quality}
\label{subsec:exp-quality}

\paragraph{Convergence within a limited budget}

TRACE converges within a small number of training iterations, reaching stable
completion and reward levels (Fig.~\ref{fig:exp-training}). Direct Finetune
converges substantially more slowly, approaching TRACE's mean reward only
later in training. Scratch remains at low completion and reward levels and
does not produce a usable tracker within the 3k-iteration budget.

\paragraph{Motion Tracking Capability}
TRACE retains a broad tracking repertoire after transfer, completing 96.08\%
of the evaluation motions, within 1.51 percentage points of SONIC/G1
(Table~\ref{tab:exp-transfer}). Local position error is 55.81\,mm on H2
versus 27.81\,mm on G1, with higher temporal errors on H2.
SR uses the height-based success criterion in~\cite{luo2025sonic}:
the absolute world-height errors of the root, wrists, and ankles must
remain within the threshold at every evaluated frame of a clip.
Scaling the G1 threshold by the whole-body
height ratio $1.820/1.320$ preserves the same error tolerance relative to
robot height.

Figure~\ref{fig:exp-tracking} shows the adapted controller executing squats,
single-knee kneels, kicks, and single-leg hops in simulation and on
H2 hardware. These motions span large posture changes, single-leg support,
and dynamic limb coordination.

\begin{figure*}[t]
  \centering
  \scalebox{1}[0.94]{%
  \begin{minipage}{0.92\textwidth}
  \includegraphics[width=\linewidth]{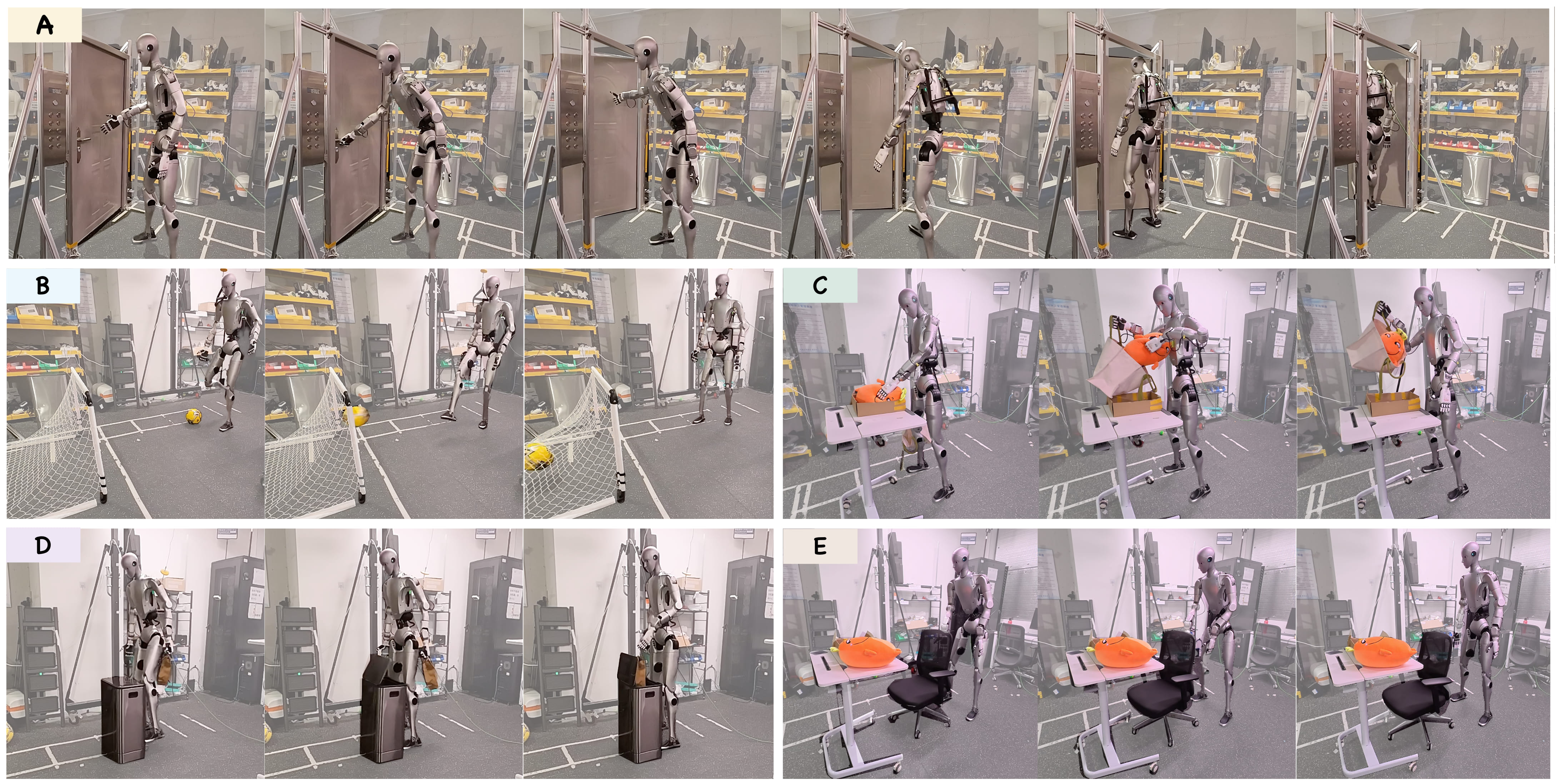}
  \end{minipage}%
  }
  \caption{Whole-body teleoperation and downstream applications on H2:
  (A)~door opening and passage, (B)~soccer, (C)~toy packing, (D)~trash disposal,
  and (E)~chair repositioning. Keyframes progress left to right, showing
  execution across objects and contact conditions.}
  \label{fig:exp-contact}
\end{figure*}

\setcounter{topnumber}{1}
\begin{table}[t]
  \centering
  \caption{Final H2 SONIC transfer training statistics, with simulation
  rollouts using the pretrained G1 checkpoint under uniform sampling as a
  reference. Position errors: cm; joint-angle error: rad.}
  \label{tab:exp-stages}
  \small\setlength{\tabcolsep}{1.5pt}
  \renewcommand{\arraystretch}{1.15}
  \begin{tabular*}{\linewidth}{@{\hspace{2pt}\extracolsep{\fill}}lcccc@{\hspace{2pt}}}
    \hline
    Metric & \begin{tabular}[c]{@{}c@{}}Single-\\phase\end{tabular} &
    \begin{tabular}[c]{@{}c@{}}Two-\\phase\end{tabular} & $\Delta$ (\%) &
    \makebox[0.17\linewidth][c]{\begin{tabular}[c]{@{}c@{}}G1\\sim rollout\end{tabular}} \\[1pt]
    \hline
    \noalign{\vskip 2pt}
    Training reward $\uparrow$ & 62.57 & 70.04 & $+11.9$ & --- \\
    Root-position error $\downarrow$ & 42.45 & 31.34 & $-26.2$ & 18.62 \\
    Global foot error $\downarrow$ & 42.94 & 31.81 & $-25.9$ & 19.04 \\
    Local hand error $\downarrow$ & 7.09 & 6.40 & $-9.7$ & 6.31 \\
    Joint-angle error $\downarrow$ & 0.12555 & 0.11957 & $-4.8$ & 0.1146 \\
    Critic value loss & 0.28871 & 0.20108 & $-30.4$ & --- \\[2pt]
    \hline
  \end{tabular*}
\end{table}

\begin{table}[t]
  \centering
  \caption{VLA task success rates on the physical H2 robot.}
  \label{tab:exp-vla}
  \small\setlength{\tabcolsep}{2pt}
  \renewcommand{\arraystretch}{1.15}
  \begin{tabular*}{\linewidth}{@{\extracolsep{\fill}}llcc@{}}
    \hline
    Task & Type & Trials & Success (\%) \\[1pt]
    \hline
    \noalign{\vskip 2pt}
    Object relocation (toy) & Single-arm & 20 & 90 \\
    Object relocation (cube) & Bimanual & 20 & 75 \\
    Object pickup (pillow) & Whole-body & 20 & 80 \\
    Open door & Whole-body & 20 & 65 \\
    Open trash can & Whole-body & 20 & 65 \\[2pt]
    \hline
  \end{tabular*}
\end{table}

\subsection{Ablation of the Adaptation Design}
\label{subsec:exp-ablation}

\paragraph{Pre-adaptation}
Kinematic mismatch and learned control corrections interact during
BFM transfer learning. We therefore assess pre-adaptation as a complete
initialization, encompassing kinematic prior injection, action-scale
alignment, and nominal joint-response alignment. This comparison evaluates
their combined effect without independently resolving each component's
contribution. Omitting pre-adaptation before target-robot training reduces
SR from 96.08\% to 87.29\% and increases $E_{\mathrm{loc}}$ from 55.81 to
72.88\,mm (Table~\ref{tab:exp-transfer}), producing the largest SR drop among
the three component ablations. These results support jointly aligning robot
interfaces and nominal responses before fine-tuning to improve the quality
of transferred whole-body control.

\paragraph{Data strategy}
The data strategy balances category frame budgets and ranks clips within
each category using frozen SONIC descriptors (Sec.~\ref{subsec:transfer-set}).
We first examine the resulting distributions by comparing uniform and
enhanced 10M-frame subsets from the same candidate pool
(Fig.~\ref{fig:exp-data-strategy}). Uniform sampling remains close to the
source pool in feature statistics, whereas enhanced selection increases the
geometric mean of feature standard deviations from 1.001 to 1.261 and yields
a higher covariance log-volume. It also increases the frame shares of
underrepresented crouching, reaching, kneeling, and balance motions,
raising normalized category entropy from 0.789 in the source pool to 0.976
in the selected set.

We then assess tracking behavior by comparing full TRACE with the variant
without the data strategy (Table~\ref{tab:exp-transfer}). Removing the
strategy reduces SR from 96.08\% to 93.80\%, a drop of 2.28 percentage
points, and increases both velocity and acceleration errors. In rollouts,
the ablated policy tracks common motions well but fails on some highly
dynamic motions that full TRACE completes. Its slightly lower
$E_{\mathrm{loc}}$ (54.01 versus 55.81\,mm) is averaged over a smaller
successful-motion set that excludes these failures; full TRACE's average
includes additional challenging executions. The SR gain and rollout
observations support the role of data selection in broadening the adapted
controller's motion repertoire.

\paragraph{Two-phase transfer learning curriculum}
We compare the two-phase curriculum with a single-phase baseline at the training
endpoint (Table~\ref{tab:exp-stages}). The curriculum starts with balanced
motion sampling, then merges the LoRA updates, reinitializes the adapters,
and introduces difficult-motion resampling.
The baseline uses the second training phase's reward terms, weights, and
other hyperparameters throughout, including difficult-motion resampling
from the outset. For matched cumulative LoRA rank budgets, the baseline
uses rank \(2r_{\mathrm{L}}=64\), versus \(r_{\mathrm{L}}=32\) per training phase.

Two-phase fine-tuning increases reward by 11.9\% and reduces root-position
and global foot errors by 26.2\% and 25.9\%, with lower local hand and
joint-angle errors (Table~\ref{tab:exp-stages}).
The H2/G1 global-error ratios are broadly in line with their body-size ratio;
G1 rollouts exclude difficult-motion resampling, favoring lower errors.
Removing the curriculum also lowers benchmark SR to 95.64\% and raises
$E_{\mathrm{loc}}$ to 62.47\,mm (Table~\ref{tab:exp-transfer}). In our runs,
early difficult-motion resampling kept policy standard deviation elevated,
supporting stable, broad tracking before emphasizing difficult motions.

\subsection{Whole-Body Control and Downstream Applications}
\label{subsec:exp-contact}

\paragraph{Whole-body interaction and teleoperation}
Figure~\ref{fig:exp-contact} shows H2 performing door opening and passage,
toy packing, trash disposal, chair repositioning, and soccer. The transferred
controller coordinates body motion, manipulation, and foot contact, while
the teleoperation interface supports task execution and demonstration collection.

\paragraph{Integration with VLA policies}
Motion generation and VLA share the same H2 low-level BFM weights and motion
interface. We post-trained GR00T-N1.7~\cite{nvidia2025groot} on selected tasks.
Across five VLA tasks with 20 trials each, success rates span 65\%--90\%
(Table~\ref{tab:exp-vla}), covering single-arm, bimanual, and whole-body
execution, including trash-can opening with foot contact.

\setcounter{topnumber}{\traceexpsavedtopnumber}

\section{Conclusion}
\label{sec:conclusion}

TRACE demonstrates that a pretrained humanoid BFM can be efficiently adapted
to a new embodiment while retaining a shared motion interface for tracking and
downstream tasks. The framework combines kinematic and nominal joint-response
pre-alignment with a compact motion set selected for quality, category balance,
and command-space dispersion. It decouples reference-encoder distillation from
feedback-control optimization within a two-phase low-rank curriculum,
establishing stable tracking before improving accuracy on difficult motions.
TRACE transfers SONIC from Unitree G1 to H2 in approximately eight hours on a
single GPU while retaining a broad motion-tracking repertoire. On physical H2,
the same adapted controller supports dynamic motion tracking, teleoperated
demonstration collection, and autonomous VLA tasks.

Our evaluation is limited to one embodiment pair and one pretrained BFM.
Future work will extend validation to more embodiments and open-source BFMs
and investigate efficient adaptation without a reusable critic. For example, multi-expert
distillation may provide no unified critic, whereas changes in training
paradigm may make existing value estimates unsuitable for the new objective.
We will also explore incorporating RGB, depth, and tactile observations
during transfer to develop perceptive BFMs.



\begin{thebibliography}{00}
\bibitem{yuan2025bfmsurvey}
M. Yuan \emph{et~al.}, ``A survey of behavior foundation model:
next-generation whole-body control system of humanoid robots,'' \emph{IEEE
Trans. Pattern Anal. Mach. Intell.}, vol.~48, no.~4, pp.~4909--4927,
Apr.~2026, doi:~10.1109/TPAMI.2025.3649177.

\bibitem{luo2025sonic}
Z. Luo \emph{et~al.}, ``SONIC: supersizing motion tracking for natural
humanoid whole-body control,'' \emph{Sci. Robot.}, vol.~11, no.~117,
Art.~no.~eaed4592, Aug.~2026, doi:~10.1126/scirobotics.aed4592.

\bibitem{lin2025hzero}
Y. Lin \emph{et~al.}, ``H-Zero: cross-humanoid locomotion pretraining enables
few-shot novel embodiment transfer,'' 2025, arXiv:2512.00971.

\bibitem{xue2026xhugwbc}
Y. Xue \emph{et~al.}, ``Scalable and general whole-body control for
cross-humanoid locomotion,'' 2026, arXiv:2602.05791.

\bibitem{peng2026eagle}
Q. Peng \emph{et~al.}, ``Embodiment-aware generalist specialist distillation
for unified humanoid whole-body control,'' 2026, arXiv:2602.02960.

\bibitem{peng2018deepmimic}
X. B. Peng, P. Abbeel, S. Levine, and M. van de Panne, ``DeepMimic:
example-guided deep reinforcement learning of physics-based character
skills,'' \emph{ACM Trans. Graph.}, vol.~37, no.~4, Art.~no.~143,
pp.~1--14, Aug.~2018, doi:~10.1145/3197517.3201311.

\bibitem{liao2026beyondmimic}
Q. Liao \emph{et~al.}, ``BeyondMimic: from motion tracking to versatile
humanoid control via guided diffusion,'' \emph{Sci. Robot.}, vol.~11,
no.~117, Art.~no.~eadx8924, Aug.~2026,
doi:~10.1126/scirobotics.adx8924.

\bibitem{he2025asap}
T. He \emph{et~al.}, ``ASAP: aligning simulation and real-world physics for
learning agile humanoid whole-body skills,'' in \emph{Proc. Robot.: Sci.
Syst. (RSS)}, 2025, doi:~10.15607/RSS.2025.XXI.066.

\bibitem{zhang2025hub}
T. Zhang \emph{et~al.}, ``HuB: learning extreme humanoid balance,'' in
\emph{Proc. 9th Conf. Robot Learn. (CoRL)}, vol.~305, pp.~686--704, 2025.

\bibitem{luo2023phc}
Z. Luo, J. Cao, A. Winkler, K. Kitani, and W. Xu, ``Perpetual humanoid
control for real-time simulated avatars,'' in \emph{Proc. IEEE/CVF Int. Conf.
Comput. Vis. (ICCV)}, 2023, pp.~10895--10904,
doi:~10.1109/ICCV51070.2023.01000.

\bibitem{he2025omnih2o}
T. He \emph{et~al.}, ``OmniH2O: universal and dexterous human-to-humanoid
whole-body teleoperation and learning,'' in \emph{Proc. 8th Conf. Robot Learn.
(CoRL)}, vol.~270, pp.~1516--1540, 2025.

\bibitem{he2025hover}
T. He \emph{et~al.}, ``HOVER: versatile neural whole-body controller for
humanoid robots,'' in \emph{Proc. IEEE Int. Conf. Robot. Autom. (ICRA)}, 2025,
pp.~9989--9996, doi:~10.1109/ICRA55743.2025.11128549.

\bibitem{li2026bfmzero}
Y. Li \emph{et~al.}, ``BFM-Zero: a promptable behavioral foundation model for
humanoid control using unsupervised reinforcement learning,'' in \emph{Proc.
Int. Conf. Learn. Represent. (ICLR)}, 2026.

\bibitem{qi2026humanoidgpt}
Z. Qi \emph{et~al.}, ``Humanoid-GPT: scaling data and structure for zero-shot
motion tracking,'' 2026, arXiv:2606.03985.

\bibitem{chen2026holomotion}
M. Chen \emph{et~al.}, ``HoloMotion-1 technical report,'' 2026,
arXiv:2605.15336.

\bibitem{gupta2022metamorph}
A. Gupta, L. Fan, S. Ganguli, and L. Fei-Fei, ``MetaMorph: learning universal
controllers with transformers,'' in \emph{Proc. Int. Conf. Learn. Represent.
(ICLR)}, 2022.

\bibitem{yang2025multiloco}
S. Yang \emph{et~al.}, ``Multi-Loco: unifying multi-embodiment legged
locomotion via reinforcement learning augmented diffusion,'' in \emph{Proc.
9th Conf. Robot Learn. (CoRL)}, vol.~305, pp.~1030--1048, 2025.

\bibitem{liu2025locoformer}
M. Liu, D. Pathak, and A. Agarwal, ``LocoFormer: generalist locomotion via
long-context adaptation,'' in \emph{Proc. 9th Conf. Robot Learn. (CoRL)},
vol.~305, pp.~532--546, 2025.

\bibitem{ha2026transformertransformer}
H. Ha, C. K. Liu, and S. Song, ``Transformer Transformer: a unified model for
motion-conditioned robot co-design,'' 2026, arXiv:2607.25798.

\bibitem{zeng2025bfm}
W. Zeng \emph{et~al.}, ``Behavior foundation model for humanoid robots,''
2025, arXiv:2509.13780.

\bibitem{wang2026fast}
Z. Wang \emph{et~al.}, ``General humanoid whole-body control via pretraining
and fast adaptation,'' 2026, arXiv:2602.11929.

\bibitem{zhao2025resmimic}
S. Zhao \emph{et~al.}, ``ResMimic: from general motion tracking to humanoid
whole-body loco-manipulation via residual learning,'' 2025, arXiv:2510.05070.

\bibitem{wang2026omnixtreme}
Y. Wang \emph{et~al.}, ``OmniXtreme: breaking the generality barrier in
high-dynamic humanoid control,'' in \emph{Proc. Robot.: Sci. Syst. (RSS)},
Sydney, Australia, Jul.~2026, doi:~10.15607/RSS.2026.XXII.031.

\bibitem{liu2022revolver}
X. Liu, D. Pathak, and K. M. Kitani, ``REvolveR: continuous evolutionary
models for robot-to-robot policy transfer,'' in \emph{Proc. 39th Int. Conf.
Mach. Learn. (ICML)}, ser.~PMLR, vol.~162, pp.~13995--14007, 2022.

\bibitem{liu2024metaevolve}
X. Liu, D. Pathak, and D. Zhao, ``Meta-Evolve: continuous robot evolution for
one-to-many policy transfer,'' in \emph{Proc. Int. Conf. Learn. Represent.
(ICLR)}, pp.~22659--22677, 2024.

\bibitem{yang2026any2any}
M. Yang, T. Yu, F. Li, and H. Chen, ``Any2Any: efficient cross-embodiment
transfer for humanoid whole-body tracking,'' 2026, arXiv:2605.23733.

\bibitem{hu2022lora}
E. J. Hu \emph{et~al.}, ``LoRA: low-rank adaptation of large language
models,'' in \emph{Proc. Int. Conf. Learn. Represent. (ICLR)}, 2022.

\bibitem{bones_seed}
Bones Studio, ``BONES-SEED: Skeletal Everyday Embodiment Dataset,''
Hugging Face, dataset. Accessed: Sep.~10, 2026. [Online]. Available:
https://huggingface.co/datasets/bones-studio/seed

\bibitem{lee2025phuma}
K. Lee \emph{et~al.}, ``PHUMA: physically reliable humanoid locomotion
dataset,'' 2025, arXiv:2510.26236.

\bibitem{nvidia2025groot}
NVIDIA \emph{et~al.}, ``GR00T N1: an open foundation model for generalist
humanoid robots,'' 2025, arXiv:2503.14734.

\end{thebibliography}
\end{document}